\documentclass[runningheads]{llncs}
\usepackage{graphicx}
\usepackage{amsmath,amssymb} 
\usepackage{color}
\usepackage{algorithm}
\usepackage{algorithmicx}
\usepackage[noend]{algpseudocode}
\usepackage{multirow}
\usepackage[width=122mm,left=12mm,paperwidth=146mm,height=193mm,top=12mm,paperheight=217mm]{geometry}

\usepackage[caption=false]{subfig}
\usepackage{caption} 

\usepackage{hyperref}
\usepackage{breakcites}

\makeatletter
\newcommand{\printfnsymbol}[1]{%
  \textsuperscript{\@fnsymbol{#1}}%
}
\makeatother

\begin{document}
\title{Mask TextSpotter: An End-to-End Trainable Neural Network for Spotting Text with Arbitrary Shapes} 

\titlerunning{Mask TextSpotter}
%
\author{Pengyuan Lyu\thanks{Authors contribute equally.}\inst{1}\orcidID{0000-0003-3153-8519
} \and
Minghui Liao\printfnsymbol{1}\inst{1}\orcidID{0000-0002-2583-4314} \and
Cong Yao\inst{2}\orcidID{0000-0001-6564-4796} \and
Wenhao Wu\inst{2} \and \\
Xiang Bai\thanks{Corresponding author.}\inst{1}\orcidID{0000-0002-3449-5940}}
%
\authorrunning{Pengyuan Lyu, Minghui Liao, Cong Yao, Wenhao Wu, Xiang Bai}
%

\institute{
Huazhong University of Science and Technology \and
Megvii (Face++) Technology Inc. \\
\email{lvpyuan@gmail.com, mhliao@hust.edu.cn, yaocong2010@gmail.com, wwh@megvii.com, xbai@hust.edu.cn}
}
\maketitle              
\begin{abstract}
Recently, models based on deep neural networks have dominated the fields of scene text detection and recognition. In this paper, we investigate the problem of scene text spotting, which aims at simultaneous text detection and recognition in natural images. An end-to-end trainable neural network model for scene text spotting is proposed. The proposed model, named as Mask TextSpotter, is inspired by the newly published work Mask R-CNN. Different from previous methods that also accomplish text spotting with end-to-end trainable deep neural networks, Mask TextSpotter takes advantage of simple and smooth end-to-end learning procedure, in which precise text detection and recognition are acquired via semantic segmentation. Moreover, it is superior to previous methods in handling text instances of irregular shapes, for example, curved text. Experiments on ICDAR2013, ICDAR2015 and Total-Text demonstrate that the proposed method achieves state-of-the-art results in both scene text detection and end-to-end text recognition tasks.

\end{abstract}

\section{Introduction}
In recent years, scene text detection and recognition have attracted growing research interests from the computer vision community, especially after the revival of neural networks and growth of image datasets. Scene text detection and recognition provide an automatic, rapid approach to access the textual information embodied in natural scenes, benefiting a variety of real-world applications, such as geo-location~\cite{zhu2018cascaded}, instant translation, and assistance for the blind.  

Scene text spotting, which aims at concurrently localizing and recognizing text from natural scenes, have been previously studied in numerous works~\cite{wang2011end,jaderberg2016reading}. However, in most works, except~\cite{Li_2017_ICCV} and~\cite{Busta_2017_ICCV}, text detection and subsequent recognition are handled separately. Text regions are first hunted from the original image by a trained detector and then fed into a recognition module. This procedure seems simple and natural, but might lead to sub-optimal performances for both detection and recognition, since these two tasks are highly correlated and complementary. On one hand, the quality of detections larges determines the accuracy of recognition; on the other hand, the results of recognition can provide feedback to help reject false positives in the phase of detection.

Recently, two methods~\cite{Li_2017_ICCV,Busta_2017_ICCV} that devise end-to-end trainable frameworks for scene text spotting have been proposed. Benefiting from the complementarity between detection and recognition, these unified models significantly outperform previous competitors. However, there are two major drawbacks in~\cite{Li_2017_ICCV} and~\cite{Busta_2017_ICCV}. First, both of them can not be completely trained in an end-to-end manner. \cite{Li_2017_ICCV} applied a curriculum learning paradigm~\cite{bengio2009curriculum} in the training period, where the sub-network for text recognition is locked at the early iterations and the training data for each period is carefully selected. Busta~\emph{et al.}~\cite{Busta_2017_ICCV} at first pre-train the networks for detection and recognition separately and then jointly train them until convergence. There are mainly two reasons that stop ~\cite{Li_2017_ICCV} and~\cite{Busta_2017_ICCV} from training the models in a smooth, end-to-end fashion. One is that the text recognition part requires accurate locations for training while the locations in the early iterations are usually inaccurate.The other is that the adopted LSTM~\cite{lstm} or CTC loss~\cite{graves2006connectionist} are difficult to optimize than general CNNs. The second limitation of~\cite{Li_2017_ICCV} and~\cite{Busta_2017_ICCV} lies in that these methods only focus on reading horizontal or oriented text. However, the shapes of text instances in real-world scenarios may vary significantly, from horizontal or oriented, to curved forms.

\begin{figure}[!b]
\begin{center}
\includegraphics[width=0.8\linewidth]{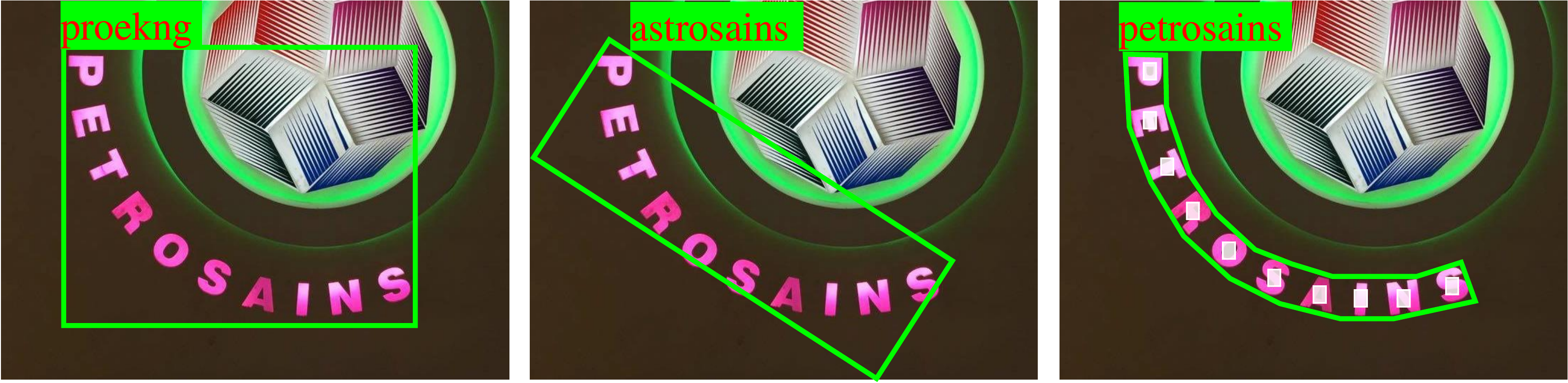}
\end{center}
\caption{Illustrations of different text spotting methods. The left presents horizontal text spotting methods~\cite{liao2017textboxes,Li_2017_ICCV}; The middle indicates oriented text spotting methods~\cite{Busta_2017_ICCV}; The right is our proposed method. Green bounding box: detection result; Red text in green background: recognition result.}
\label{fig:introduction}
\end{figure}

In this paper, we propose a text spotter named as \emph{Mask TextSpotter}, which can detect and recognize text instances of arbitrary shapes. Here, \textit{arbitrary shapes} mean various forms text instances in real world. Inspired by Mask R-CNN~\cite{he2017mask}, which can generate shape masks of objects, we detect text by segment the instance text regions. Thus our detector is able to detect text of arbitrary shapes. Besides, different from the previous sequence-based recognition methods \cite{shi2016robust,shi2017end,lee2016recursive} which are designed for 1-D sequence, we recognize text via semantic segmentation in 2-D space, to solve the issues in reading irregular text instances. Another advantage is that it does not require accurate locations for recognition. Therefore, the detection task and recognition task can be completely trained end-to-end, and benefited from feature sharing and joint optimization.

We validate the effectiveness of our model on the datasets that include horizontal, oriented and curved text. The results demonstrate the advantages of the proposed algorithm in both text detection and end-to-end text recognition tasks. Specially, on ICDAR2015, evaluated at a single scale, our method achieves an F-Measure of 0.86 on the detection task and outperforms the previous top performers by $13.2\%-25.3\%$ on the end-to-end recognition task.

The main contributions of this paper are four-fold. (1) We propose an end-to-end trainable model for text spotting, which enjoys a simple, smooth training scheme. (2) The proposed method can detect and recognize text of various shapes, including horizontal, oriented, and curved text. (3) In contrast to previous methods, precise text detection and recognition in our method are accomplished via semantic segmentation. (4) Our method achieves state-of-the-art performances in both text detection and text spotting on various benchmarks.

\section{Related Work}

\subsection{Scene Text Detection}
In scene text recognition systems, text detection plays an important role~\cite{zhu2016scene}. A large number of methods have been proposed to detect scene text \cite{epshtein2010detecting,neumann2010method,neumann2012real,yao2012detecting,huang2014robust,kang2014orientation,zhang2015symmetry,jaderberg2016reading,tian2015text,zhang2015symmetry,zhong2016deeptext,liao2017textboxes,yao2016scene,zhang2016multi,liu2017deep,he2017single,tian2016detecting,shi2017detecting,zhou2017east,he2017deep,lyu2018multi,liao2018rotation}. 
In \cite{jaderberg2016reading}, Jaderberg \emph{et al.} use Edge Boxes \cite{zitnick2014edge} to generate proposals and refine candidate boxes by regression. 
Zhang \emph{et al.} \cite{zhang2015symmetry} detect scene text by exploiting the symmetry property of text. 
Adapted from Faster R-CNN \cite{ren2015faster} and SSD \cite{liu2016ssd} with well-designed modifications, \cite{zhong2016deeptext,liao2017textboxes} are proposed to detect horizontal words.

Multi-oriented scene text detection has become a hot topic recently. Yao \emph{et al.} \cite{yao2016scene} and Zhang \emph{et al.} \cite{zhang2016multi} detect multi-oriented scene text by semantic segmentation. 
Tian \emph{et al.} \cite{tian2016detecting} and Shi \emph{et al.} \cite{shi2017detecting} propose methods which first detect text segments and then link  them into text instances by spatial relationship or link predictions.
Zhou \emph{et al.} \cite{zhou2017east} and He \emph{et al.} \cite{he2017deep} regress text boxes directly from dense segmentation maps. 
Lyu \emph{et al.} \cite{lyu2018multi} propose to detect and group the corner points of the text to generate text boxes. Rotation-sensitive regression for oriented scene text detection is proposed by Liao \emph{et al.} \cite{liao2018rotation}.

Compared to the popularity of horizontal or multi-oriented scene text detection, there are few works focusing on text instances of arbitrary shapes. 
Recently, detection of text with arbitrary shapes has gradually drawn the attention of researchers due to the application requirements in the real-life scenario. In \cite{risnumawan2014robust}, Risnumawan \emph{et al.} propose a system for arbitrary text detection based on text symmetry properties. In~\cite{CK2017}, a dataset which focuses on curve orientation text detection is proposed. Different from most of the above-mentioned methods, we propose to detect scene text by instance segmentation which can detect text with arbitrary shapes.

\subsection{Scene Text Recognition}

Scene text recognition~\cite{yao2014strokelets,shi2018aster} aims at decoding the detected or cropped image regions into character sequences. The previous scene text recognition approaches can be roughly split into three branches: character-based methods, word-based methods, and sequence-based methods. The character-based recognition methods \cite{bissacco2013photoocr,jaderberg2014deep} mostly first localize individual characters and then recognize and group them into words. In \cite{Jaderberg14c}, Jaderberg \emph{et al.} propose a word-based method which treats text recognition as a common English words (90k) classification problem. Sequence-based methods solve text recognition as a sequence labeling problem. In \cite{shi2017end}, Shi \emph{et al.} use CNN and RNN to model image features and output the recognized sequences with CTC \cite{graves2006connectionist}. In \cite{lee2016recursive,shi2016robust}, Lee \emph{et al.} and Shi \emph{et al.} recognize scene text via attention based sequence-to-sequence model.

The proposed text recognition component in our framework can be classified as a character-based method. However, in contrast to previous character-based approaches, we use an FCN \cite{long2015fully} to localize and classify characters simultaneously. Besides, compared with sequence-based methods which are designed for a 1-D sequence, our method is more suitable to handle irregular text (multi-oriented text, curved text \emph{et al.}).

\subsection{Scene Text Spotting}

Most of the previous text spotting methods \cite{jaderberg2016reading,liao2017textboxes,SynthText,liao2018textboxes++} split the spotting process into two stages. They first use a scene text detector \cite{jaderberg2016reading,liao2017textboxes,liao2018textboxes++} to localize text instances and then use a text recognizer \cite{Jaderberg14c,shi2017end} to obtain the recognized text. In \cite{Li_2017_ICCV,Busta_2017_ICCV}, Li \emph{et al.} and Busta \emph{et al.} propose end-to-end methods to localize and recognize text in a unified network, but require relatively complex training procedures. Compared with these methods, our proposed text spotter can not only be trained end-to-end completely, but also has the ability to detect and recognize  arbitrary-shape (horizontal, oriented, and curved) scene text.

\subsection{General Object Detection and Semantic Segmentation}

With the rise of deep learning, general object detection and semantic segmentation have achieved great development. A large number of object detection and segmentation  methods \cite{girshick2014rich,fastrcnn,ren2015faster,dai2016r,lin2017feature,liu2016ssd,redmon2016you,long2015fully,dai2016instance,li2017fully,he2017mask} have been proposed. Benefited from those methods, scene text detection and recognition have achieved obvious progress in the past few years. Our method is also inspired by those methods. Specifically, our method is adapted from a general object instance segmentation model Mask R-CNN~\cite{he2017mask}. However, there are key differences between the mask branch of our method and that in Mask R-CNN. Our mask branch can not only segment text regions but also predict character probability maps, which means that our method can be used to recognize the instance sequence inside character maps rather than predicting an object mask only.

\section{Methodology}
\begin{figure}[!b]
\begin{center}
\includegraphics[width=0.75\linewidth]{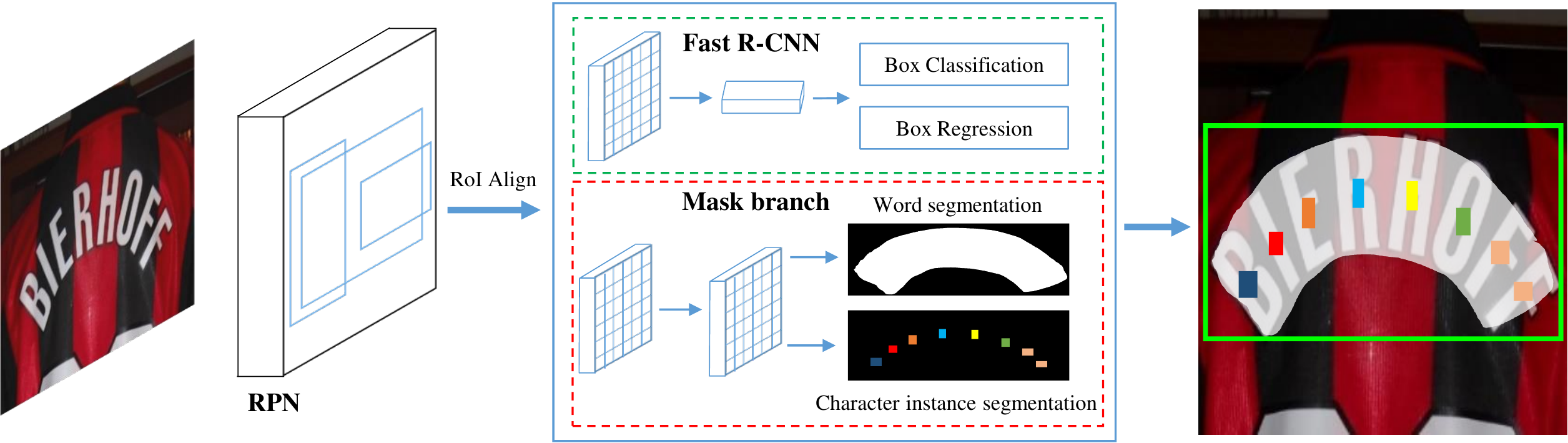}
\end{center}
\caption{Illustration of the architecture of the our method.}
\label{fig:pipeline}
\end{figure}

The proposed method is an end-to-end trainable text spotter, which can handle various shapes of text. It consists of an instance-segmentation based text detector and a character-segmentation based text recognizer.
\subsection{Framework}\label{sec:framework}

The overall architecture of our proposed method is presented in Fig.~\ref{fig:pipeline}. Functionally, the framework consists of four components: a feature pyramid network (FPN)~\cite{lin2017feature} as backbone, a region proposal network (RPN)~\cite{ren2015faster} for generating text proposals, a Fast R-CNN~\cite{ren2015faster} for bounding boxes regression, a mask branch for text instance segmentation and character segmentation. 
In the training phase, a lot of text proposals are first generated by RPN, and then the RoI features of the proposals are fed into the Fast R-CNN branch and the mask branch to generate the accurate text candidate boxes, the text instance segmentation maps, and the character segmentation maps.

\noindent\textbf{Backbone} Text in nature images are various in sizes. In order to build high-level semantic feature maps at all scales, we apply a feature pyramid structure~\cite{lin2017feature} backbone with ResNet~\cite{resnet} of depth 50.
FPN uses a top-down architecture to fuse the feature of different resolutions from a single-scale input,  which improves accuracy with marginal cost.
\\
\textbf{RPN}
RPN is used to generate text proposals for the subsequent Fast R-CNN and mask branch. Following ~\cite{lin2017feature}, we assign anchors on different stages depending on the anchor size. Specifically, the area of the anchors are set to \{$32^2, 64^2, 128^2, 256^2, 512^2$\} pixels on five stages \{$P_2, P_3, P_4, P_5, P_6$\} respectively. Different aspect ratios \{$0.5,1,2$\} are also adopted in each stages as in ~\cite{ren2015faster}. In this way, the RPN can handle text of various sizes and aspect ratios. RoI Align~\cite{he2017mask} is adapted to extract the region features of the proposals. Compared to RoI Pooling~\cite{fastrcnn}, RoI Align preserves more accurate location information, which is quite beneficial to the segmentation task in the mask branch. Note that no special design for text is adopted, such as the special aspect ratios or orientations of anchors for text, as in previous works~\cite{liao2017textboxes,he2017single,liu2017deep}.

\noindent\textbf{Fast R-CNN}
The Fast R-CNN branch includes a classification task and a regression task. The main function of this branch is to provide more accurate bounding boxes for detection. The inputs of Fast R-CNN are in $7 \times 7$ resolution, which are generated by RoI Align from the proposals produced by RPN.

\begin{figure}[!b]
\begin{center}
\includegraphics[width=0.8\linewidth]{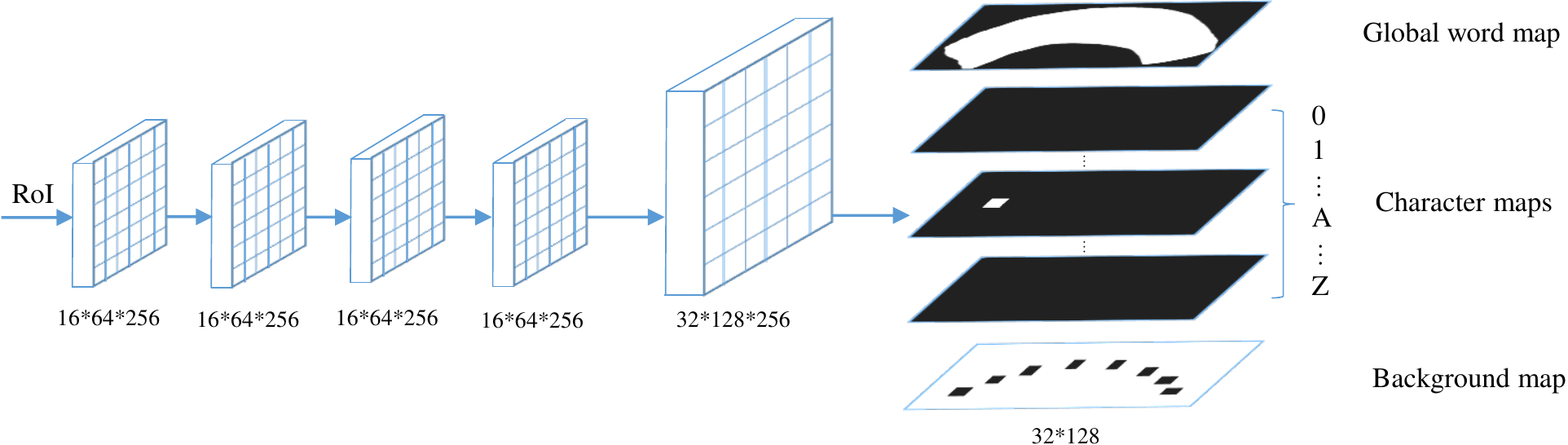}
\end{center}
\caption{Illustration of the mask branch. Subsequently, there are four convolutional layers, one de-convolutional layer, and a final convolutional layer which predicts maps of 38 channels (1 for global text instance map; 36 for character maps; 1 for background map of characters).}
\label{fig:mask_branch}
\end{figure}

\noindent\textbf{Mask Branch}
There are two tasks in the mask branch, including a global text instance segmentation task and a character segmentation task. As shown in Fig.~\ref{fig:mask_branch}, giving an input RoI, whose size is fixed to $16*64$, through four convolutional layers and a de-convolutional layer, the mask branch predicts 38 maps (with $32*128$ size), including a global text instance map, 36 character maps, and a background map of characters. 
The global text  instance map can give accurate localization of a text region, regardless of the shape of the text instance.
The character maps are maps of 36 characters, including 26 letters and 10 Arabic numerals. The background map of characters, which excludes the character regions, is also needed for post-processing.

\subsection{Label Generation}
For a training sample with the input image $I$ and the corresponding ground truth,  we generate targets for RPN, Fast R-CNN and mask branch. Generally, the ground truth contains $P=\{p_{1}, p_{2}...p_{m} \}$ and $C=\{c_{1}=(cc_{1},cl_{1}),c_{2}=(cc_{2},cl_{2}), ... , c_{n}=(cc_{n},cl_{n})\}$, where $p_{i}$ is a polygon which represents the localization of a text region, $cc_{j}$ and $cl_{j}$ are the category and location of a character respectively. Note that, in our method $C$ is not necessary for all training samples. 


\begin{figure}[!b]
\begin{center}
\captionsetup[subfigure]{justification=centering}
    \centering
\subfloat[\label{fig:label_generation}]{%
       \includegraphics[width=0.4\textwidth]{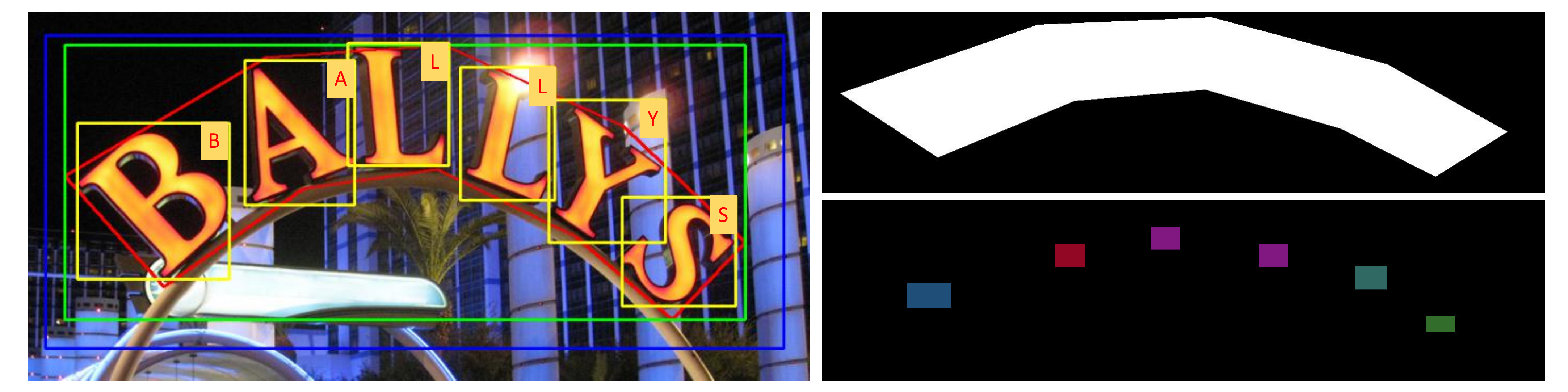}
     }
\subfloat[\label{fig:pixel_voting}]{%
       \includegraphics[width=0.5\textwidth]{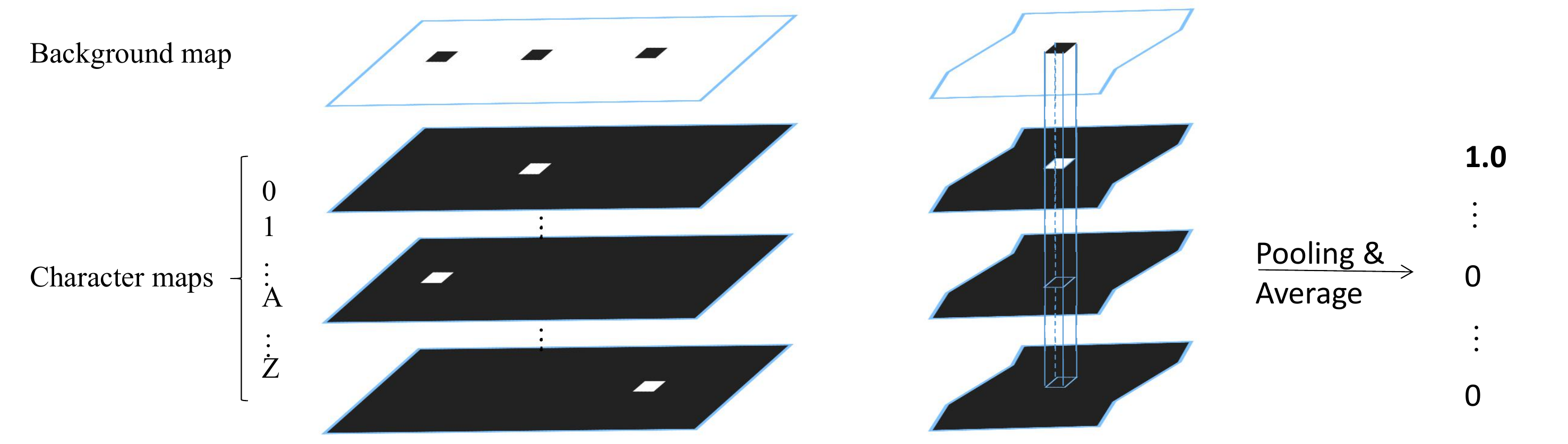}
     }
\end{center}
 \caption{(a) Label generation of mask branch. Left: the blue box is a proposal yielded by RPN, the red polygon and yellow boxes are ground truth polygon and character boxes, the green box is the horizontal rectangle which covers the polygon with minimal area. Right: the global map (top) and the character map (bottom). (b) Overview of the pixel voting algorithm. Left: the predicted character maps; right: for each connected regions, we calculate the scores for each character by averaging the probability values in the corresponding region.}
\label{fig:denser-box}
\end{figure}

We first transform the polygons into horizontal rectangles which cover the polygons with minimal areas. And then we generate targets for RPN and Fast R-CNN following \cite{fastrcnn,ren2015faster,lin2017feature}. There are two types of target maps to be generated for the mask branch with the ground truth $P$, $C$ (may not exist) as well as the proposals yielded by RPN: a global map for text instance segmentation and a character map for character semantic segmentation. 
Given a positive  proposal $r$, we first use the matching mechanism of \cite{fastrcnn,ren2015faster,lin2017feature} to obtain the best matched horizontal rectangle. The corresponding polygon as well as characters (if any) can be obtained further. Next, the matched polygon and character boxes are shifted and resized to align the proposal and the target map of $H\times W$ as  the following formulas:

\begin{equation}
B_{x}=(B_{x_0}-min(r_{x})) \times W / (max(r_{x})-min(r_{x}))
\end{equation}
\begin{equation}
B_{y}=(B_{y_0}-min(r_{y})) \times H / (max(r_{y})-min(r_{y}))
\end{equation}
where $(B_{x},B_{y})$ and $(B_{x_0},B_{y_0})$ are the updated and original vertexes of the polygon and all character boxes; $(r_{x}, r_{y})$ are the vertexes of the proposal $r$.

After that, the target global map can be generated by just drawing the normalized polygon on a zero-initialized mask and filling the polygon region with the value $1$. 
The character map generation is visualized in Fig.~\ref{fig:label_generation}. We first shrink all character bounding boxes by fixing their center point and shortening the sides to the fourth of the original sides. Then, the values of the pixels in the shrunk character bounding boxes are set to their corresponding category indices and those outside the shrunk character bounding boxes are set to $0$. If there are no character bounding boxes annotations, all values are set to $-1$.

\subsection{Optimization}
As discussed in Sec.~\ref{sec:framework}, our model includes multiple tasks. We naturally define a multi-task loss function:
\begin{equation}
L = L_{rpn} + \alpha_1 L_{rcnn} + \alpha_2 L_{mask},
\end{equation}
where $L_{rpn}$ and $L_{rcnn}$ are the loss functions of RPN and Fast R-CNN, which are identical as these in~\cite{ren2015faster} and~\cite{fastrcnn}. 
The mask loss $L_{mask}$ consists of a global text instance segmentation loss $L_{global}$ and a character segmentation loss $L_{char}$:
\begin{equation}
L_{mask} = L_{global} + \beta L_{char},
\end{equation}
where $L_{global}$ is an average binary cross-entropy loss and $L_{char}$ is a weighted spatial soft-max loss. In this work, the $\alpha_1$, $\alpha_2$, $\beta$, are empirically set to $1.0$.

\subsubsection{Text instance segmentation loss}
The output of the text instance segmentation task is a single map. Let $N$ be the number of pixels in the global map, $y_n$ be the pixel label ($y_n \in {0,1}$), and $x_n$ be the output pixel, we define the $L_{global}$ as follows:
\begin{equation}
L_{global} = -\frac{1}{N}\sum_{n=1}^{N}\left [ y_n \times log(S(x_n)) + (1-y_n) \times log(1- S(x_n)) \right ]
\end{equation}
where $S(x)$ is a sigmoid function.

\subsubsection{Character segmentation loss}
The output of the character segmentation consists of 37 maps, which correspond to 37 classes (36 classes of characters and the background class). Let $T$ be the number of classes, $N$ be the number of pixels in each map. The output maps X can be viewed as an $N \times T$ matrix. In this way, the weighted spatial soft-max loss can be defined as follows:

\begin{equation}
L_{char} = -\frac{1}{N}\sum_{n=1}^{N}W_n\sum_{t=0}^{T-1} Y_{n,t} log(\frac{e^{X_{n,t}}}{\sum_{k=0}^{T-1} e^{X_{n,k}}}),
\end{equation}

where $Y$ is the corresponding ground truth of $X$. The weight $W$ is used to balance the loss value of the positives (character classes) and the background class. Let the number of the background pixels be $N_{neg}$, and the background class index be $0$, the weights can be calculated as:

\begin{equation}
  W_i = 
  \begin{cases}
    1& \text{if } Y_{i,0}=1, \\
    N_{neg} / (N - N_{neg})& \text{otherwise}
  \end{cases}
\end{equation}

Note that in inference, a sigmoid function and a soft-max function are applied to generate the global map and the character segmentation maps respectively.

\subsection{Inference}

Different from the training process where the input RoIs of mask branch come from RPN, in the inference phase, we use the outputs of Fast R-CNN as proposals to generate the predicted global maps and character maps, since the Fast R-CNN outputs are more accurate. 

Specially, the processes of inference are as follows: first, inputting a test image, we obtain the outputs of Fast R-CNN as \cite{ren2015faster} and filter out the redundant candidate boxes by NMS; and then, the kept proposals are fed into the mask branch to generate the global maps and the character maps; finally the predicted polygons can be obtained directly by calculating the contours of text regions on global maps, the character sequences can be generated by our proposed \textit{pixel voting} algorithm on character maps.

\noindent\textbf{Pixel Voting} We decode the predicted character maps into character sequences by our proposed pixel voting algorithm. We first binarize the background map, where the values are from $0$ to $255$, with a threshold of $192$. 
Then we obtain all character regions according to connected regions in the binarized map. We calculate the mean values of each region for all character maps. The values can be seen as the character classes probability of the region. The character class with the largest mean value will be assigned to the region. After that, we group all the characters from left to right according to the writing habit of English. 

\noindent\textbf{Weighted Edit Distance} Edit distance can be used to find the best-matched word of a predicted sequence with a given lexicon. However, there may be multiple words matched with the minimal edit distance at the same time, and the algorithm can not decide which one is the best. The main reason for the above-mentioned issue is that all operations (delete, insert, replace) in the original edit distance algorithm have the same costs, which does not make sense actually. 

\begin{figure}[!bp]
\begin{center}
\includegraphics[width=0.8\linewidth]{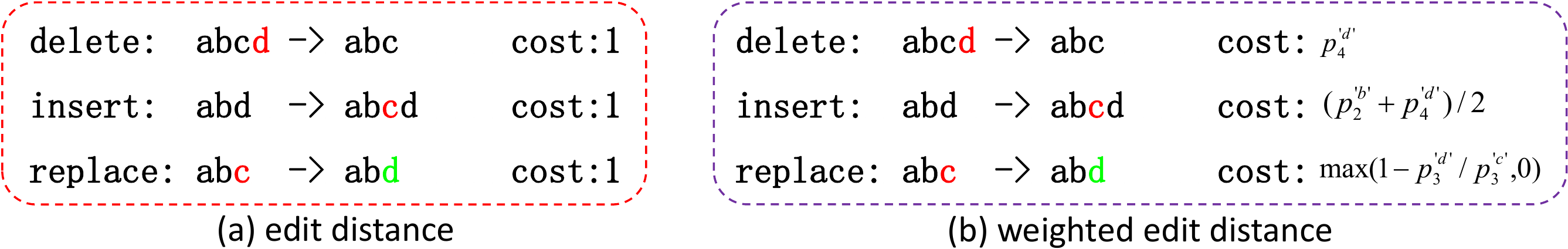}
\end{center}
\caption{Illustration of the edit distance and our proposed weighted edit distance. The red characters are the characters will be deleted, inserted and replaced. Green characters mean the candidate characters. $p_{index}^c$ is the character probability, $index$ is the character index and $c$ is the current character.}
\label{fig:wed}
\end{figure}

Inspired by~\cite{yao2014unified}, we propose a weighted edit distance algorithm. As shown in Fig.~\ref{fig:wed}, different from edit distance,  which assign the same cost for different operations, the costs of our proposed weighted edit distance depend on the character probability $p_{index}^c$ which yielded by the pixel voting. Mathematically, the weighted edit distance between two strings $a$ and $b$, whose length are $|a|$ and $|b|$ respectively, can be described as $D_{a,b}(|a|,|b|)$, where
\begin{equation}
{\qquad D_{a,b}(i,j)={\begin{cases}\max(i,j)&{\text{ if }}\min(i,j)=0,\\\min {\begin{cases} D_{a,b}(i-1,j)+C_d\\ D_{a,b}(i,j-1)+C_i\\ D_{a,b}(i-1,j-1)+C_r \times 1_{(a_{i}\neq b_{j})}\end{cases}}&{\text{ otherwise.}}\end{cases}}}
\end{equation}
where $1_{(a_{i}\neq b_{j})}$ is the indicator function equal to 0 when $a_{i}=b_{j}$ and equal to 1 otherwise; $D_{a,b}(i,j)$ is the distance between the first $i$ characters of $a$ and the first $j$ characters of $b$; $C_d$, $C_i$, and $C_r$ are the deletion, insert, and replace cost respectively. In contrast, these costs are set to $1$ in the standard edit distance.

\section{Experiments}

To validate the effectiveness of the proposed method, we conduct experiments and compare with other state-of-the-art methods on three public datasets: a horizontal text set ICDAR2013 \cite{karatzas2013icdar}, an oriented text set ICDAR2015 \cite{karatzas2015icdar} and a curved text set Total-Text \cite{CK2017}.

\subsection{Datasets}
\textbf{SynthText} is a synthetic dataset proposed by \cite{SynthText}, including about 800000 images. Most of the text instances in this dataset are multi-oriented and annotated  with word and character-level rotated bounding boxes, as well as text sequences.
\\
\textbf{ICDAR2013} is a dataset proposed in Challenge 2 of the ICDAR 2013 Robust Reading Competition \cite{karatzas2013icdar} which focuses on the horizontal text detection and recognition in natural images. There are 229 images in the training set and 233 images in the test set. Besides, the bounding box and the transcription are also provided for each word-level and character-level text instance. 
\\
\textbf{ICDAR2015} is proposed in Challenge 4 of the ICDAR 2015 Robust Reading Competition \cite{karatzas2015icdar}. Compared to ICDAR2013 which focuses on ``focused text" in particular scenario, ICDAR2015 is more concerned with the incidental scene text detection and recognition. It contains 1000 training samples and 500 test images. All training images are annotated with word-level quadrangles as well as corresponding transcriptions. Note that, only localization annotations of words are used in our training stage.
\\
\textbf{Total-Text} is a comprehensive scene text dataset proposed by \cite{CK2017}. Except for the horizontal text and oriented text, Total-Text also consists of a lot of curved text. Total-Text contains 1255 training images and 300 test images. All images are annotated with polygons and transcriptions in word-level. Note that, we only use the localization annotations in the training phase.

\subsection{Implementation details}
\noindent\textbf{Training} Different from previous text spotting methods which use two independent models \cite{jaderberg2014deep,liao2017textboxes} (the detector and the recognizer) or alternating training strategy \cite{Li_2017_ICCV}, all subnets of our model can be trained synchronously and end-to-end. The whole training process contains two stages: pre-trained on SynthText and fine-tuned on the real-world data. 

In the pre-training stage, we set the mini-batch to $8$, and all the shorter edge of the input images are resized to $800$ pixels while keeping the aspect ratio of the images. The batch sizes of RPN and Fast R-CNN are set to $256$ and $512$ per image with a $1:3$ sample ratio of positives to negatives. The batch size of the mask branch is $16$. In the fine-tuning stage, data augmentation and multi-scale training technology are applied due to the lack of real samples. Specifically, for data augmentation, we randomly rotate the input pictures in a certain angle range of $[-15^\circ, 15^\circ]$. Some other augmentation tricks, such as modifying the hue, brightness, contrast randomly, are also used following \cite{liu2016ssd}. For multi-scale training, the shorter sides of the input images are randomly resized to three scales (600, 800, 1000). Besides, following \cite{Li_2017_ICCV}, extra 1162 images for character detection from \cite{zhong2016deeptext} are also used as training samples. The mini-batch of images is kept to 8, and in each mini-batch, the sample ratio of different datasets is set to $4:1:1:1:1$ for SynthText, ICDAR2013, ICDAR2015, Total-Text and the extra images respectively. The batch sizes of RPN and Fast R-CNN are kept as the pre-training stage, and that of the mask branch is set to $64$ when fine-tuning.

We optimize our model using SGD with a weight decay of 0.0001 and momentum of 0.9. In the pre-training stage, we train our model for 170k iterations, with an initial learning rate of 0.005. Then the learning rate is decayed to a tenth at the 120k iteration. In the fine-tuning stage, the initial learning rate is set to 0.001, and then be decreased to 0.0001 at the 40k iteration. The fine-tuning process is terminated at the 80k iteration.

\textbf{Inference} In the inference stage, the scales of the input images depend on different datasets. After NMS, 1000 proposals are fed into Fast R-CNN. False alarms and redundant candidate boxes are filtered out by Fast R-CNN and NMS respectively. The kept candidate boxes are input to the mask branch to generate the global text instance maps and the character maps. Finally, the text instance bounding boxes and sequences are generated from the predicted maps.

We implement our method in Caffe2 and conduct all experiments on a regular workstation with Nvidia Titan Xp GPUs. The model is trained in parallel and evaluated on a single GPU.

\subsection{Horizontal text}
We evaluate our model on ICDAR2013 dataset to verify its effectiveness in detecting and recognizing horizontal text. We resize the shorter sides of all input images to 1000 and evaluate the results on-line.

The results of our model are listed and compared with other state-of-the-art methods in Table~\ref{tab_icdar2013} and Table~\ref{tab_detection}. As shown, our method achieves state-of-the-art results among detection, word spotting and end-to-end recognition. Specifically, for detection, though evaluated at a single scale, our method outperforms some previous methods which are evaluated at multi-scale setting \cite{hu2017wordsup,he2017deep} (F-Measure: $91.7\%$  \emph{v.s.}  $90.3\%$); for word spotting, our method is  comparable to the previous best method; for end-to-end recognition,  despite  amazing results have been achieved by \cite{liao2017textboxes,Li_2017_ICCV}, our method is still beyond them by $1.1\%-1.9\%$.

\subsection{Oriented text}
We verify the superiority of our method in detecting and recognizing oriented text by conducting experiments on ICDAR2015. We input the images with three different scales: the original scale ($720 \times 1280$) and two larger scales where shorter sides of the input images are 1000 and 1600 due to a lot of small text instance in ICDAR2015.  We evaluate our method on-line and compare it  with other methods in Table~\ref{tab_icdar2015} and Table~\ref{tab_detection}. Our method outperforms the previous methods by a large margin both in detection and recognition. For detection, when evaluated at the original scale, our method achieves the F-Measure of $84\%$, higher than the current best one \cite{he2017deep}  by $3.0\%$, which evaluated at multiple scales. When evaluated at a larger scale, a more impressive result can be achieved (F-Measure: $86.0\%$), outperforming the competitors by at least $5.0\%$. Besides, our method also achieves remarkable results on word spotting and end-to-end recognition. Compared with the state of the art, the performance of our method has significant improvements by $13.2\%-25.3\%$, for all evaluation situations.
\begin{table}[ht]
\begin{centering}
\caption{Results on ICDAR2013. ``S", ``W" and ``G" mean recognition with strong, weak and generic lexicon respectively.}
\label{tab_icdar2013}
\begin{tabular}{|c|c|c|c|c|c|c|c|}
\hline 
\multirow{2}{*}{Method} & \multicolumn{3}{c|}{Word Spotting} & \multicolumn{3}{c|}{End-to-End} & \multirow{2}{*}{FPS}\tabularnewline
\cline{2-7} 
 & S & W & G & S & W & G & \tabularnewline
\hline 
\hline
Jaderberg \emph{et al.} \cite{jaderberg2016reading} & 90.5  & - & 76 & 86.4 & - & - & - \tabularnewline
\hline 
 FCRNall+multi-filt \cite{SynthText} & - & - & 84.7 & - & - & - &  - \tabularnewline
 \hline 
Textboxes \cite{liao2017textboxes}  & 93.9  & 92.0  & 85.9 & 91.6 & 89.7 & 83.9 & - \tabularnewline
 \hline 
Deep text spotter \cite{Busta_2017_ICCV} & 92 & 89 & 81 & 89 & 86 & 77 & \textbf{9} \tabularnewline
\hline 
Li \emph{et al.} \cite{Li_2017_ICCV}  &\textbf{94.2} &\textbf{92.4} &\textbf{88.2} &91.1 &89.8 &84.6 & 1.1 \tabularnewline
\hline 
\hline
 Ours  &92.5 &92.0 &\textbf{88.2} &\textbf{92.2} &\textbf{91.1} &\textbf{86.5} &4.8   \tabularnewline
\hline
\end{tabular}
\par\end{centering}
\end{table}

\begin{table}[ht]
\begin{centering}
\caption{Results on ICDAR2015. ``S", ``W" and ``G" mean recognition with strong, weak and generic lexicon respectively.}
\label{tab_icdar2015}
\begin{tabular}{|c|c|c|c|c|c|c|c|}
\hline 
\multirow{2}{*}{Method} & \multicolumn{3}{c|}{Word Spotting} & \multicolumn{3}{c|}{End-to-End} & \multirow{2}{*}{FPS}\tabularnewline
\cline{2-7} 
 & S & W & G & S & W & G & \tabularnewline
\hline 
\hline
Baseline OpenCV3.0 + Tesseract\cite{karatzas2015icdar}  & 14.7 & 12.6 & 8.4 & 13.8  & 12.0  & 8.0 & - \tabularnewline
\hline 
TextSpotter \cite{neumann2016real} & 37.0  & 21.0 & 16.0 & 35.0 & 20.0 & 16.0 & 1 \tabularnewline
\hline 
Stradvision \cite{karatzas2015icdar} & 45.9  & - & - & 43.7 & - & - & - \tabularnewline
\hline 
TextProposals + DictNet \cite{gomez2017textproposals,Jaderberg14c} & 56.0 & 52.3 & 49.7 & 53.3 & 49.6 & 47.2 &  0.2 \tabularnewline
\hline 
HUST\_MCLAB \cite{shi2017detecting,shi2017end} & 70.6 & - & - & 67.9 & - & - &  - \tabularnewline
 \hline 
Deep text spotter \cite{Busta_2017_ICCV} & 58.0 & 53.0 & 51.0 & 54.0 & 51.0 & 47.0 & \textbf{9.0} \tabularnewline
\hline 
\hline
 Ours (720)  &71.6  &63.9  &51.6  &71.3  &62.5  &50.0  &6.9   \tabularnewline
\hline
 Ours (1000)  &77.7  &71.3  &58.6  &77.3  &69.9  &60.3  &4.8   \tabularnewline
\hline
 Ours (1600)  &\textbf{79.3}  &\textbf{74.5}  &\textbf{64.2}  &\textbf{79.3}  &\textbf{73.0}  &\textbf{62.4}  &2.6   \tabularnewline
\hline
\end{tabular}
\par\end{centering}
\end{table}

\begin{figure}[!tbp]
\begin{center}
\includegraphics[width=0.8\linewidth]{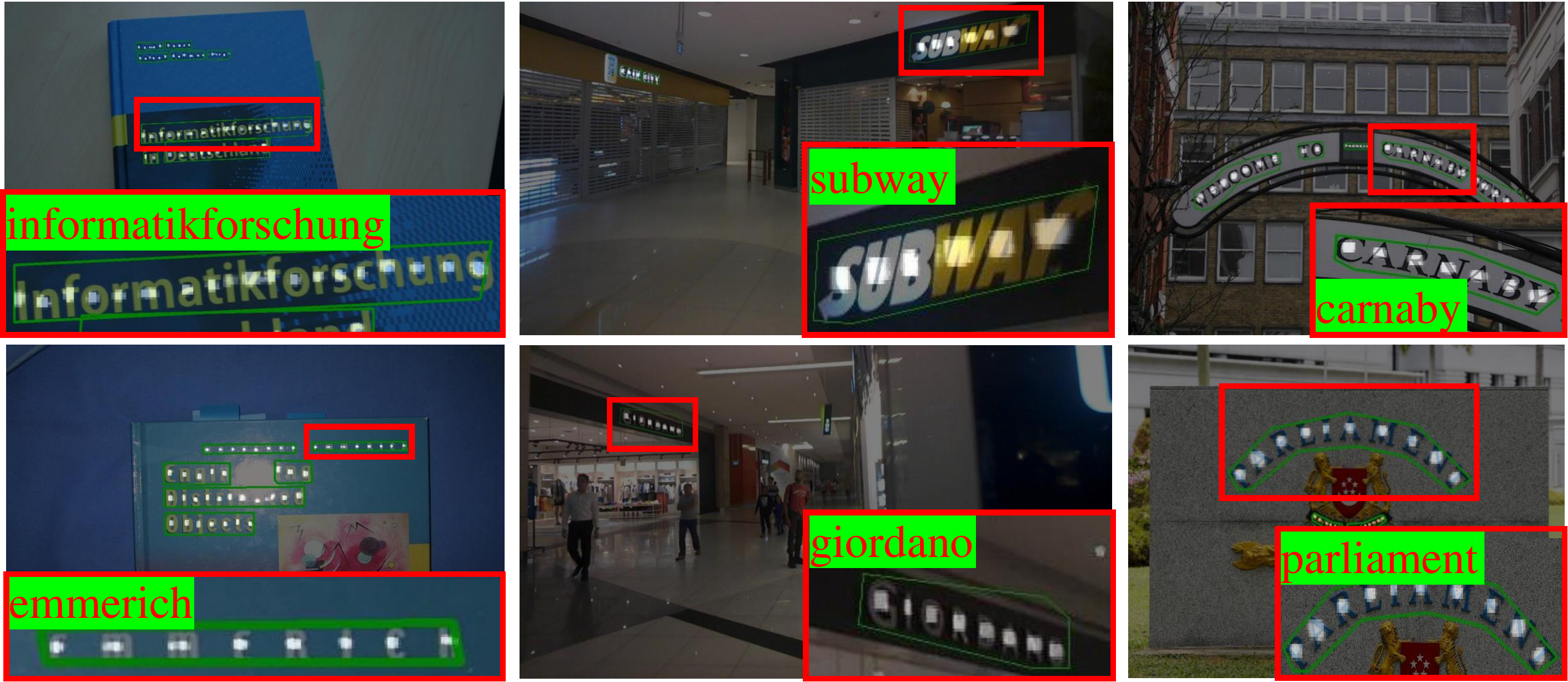}
\end{center}
\caption{Visualization results of ICDAR 2013 (the left), ICDAR 2015 (the middle) and Total-Text (the right).}
\label{fig:visu}
\end{figure}

\begin{table}[ht]
\begin{centering}
\caption{The detection results on ICDAR2013 and ICDAR2015. For ICDAR2013, all methods are evaluated  under the ``DetEval” evaluation protocol. The short sides of the input image in ``Ours (det only)" and ``Ours'' are set to $1000$.}
\label{tab_detection}
\begin{tabular}{|c|c|c|c|c|c|c|c|c|}
\hline 
\multirow{2}{*}{Method} & \multicolumn{3}{c|}{ICDAR2013} & \multirow{2}{*}{FPS} & \multicolumn{3}{c|}{ICDAR2015} & \multirow{2}{*}{FPS}\tabularnewline
\cline{2-4} \cline{6-8} 
 & Precision & Recall & F-Measure &  & Precision & Recall & F-Measure & \tabularnewline
\hline 
\hline
Zhang \emph{et al.} \cite{zhang2016multi} & 88.0 & 78.0 & 83.0 & 0.5 & 71.0 & 43.0 &  54.0 & 0.5 \tabularnewline
\hline 
Yao \emph{et al.} \cite{yao2016scene} & 88.9 & 80.2 & 84.3 & 1.6 & 72.3 & 58.7 & 64.8 & 1.6  \tabularnewline
\hline 
CTPN \cite{tian2016detecting} & 93.0 & 83.0  & 88.0 & 7.1 & 74.0 & 52.0 & 61.0 & -  \tabularnewline
\hline 
Seglink \cite{shi2017detecting} & 87.7 & 83.0 & 85.3  & \textbf{20.6} & 73.1 & 76.8 & 75.0 & - \tabularnewline
\hline 
EAST \cite{zhou2017east} & - & - & - & - & 83.3 & 78.3 & 80.7 & -   \tabularnewline
\hline 
SSTD \cite{he2017single} & 89.0 & 86.0  & 88.0 & 7.7 & 80.0 & 73.0 &  77.0 & \textbf{7.7} \tabularnewline
\hline 
Wordsup \cite{hu2017wordsup} & 93.3 & 87.5 & 90.3 & 2 & 79.3 & 77.0 & 78.2 & 2  \tabularnewline
\hline 
He \emph{et al.} \cite{he2017deep} & 92.0 & 81.0 & 86.0 & 1.1 & 82.0 & 80.0 & 81.0 & 1.1 \tabularnewline
\hline 
\hline
Ours (det only) &94.1  &88.1  &91.0  &4.6   &85.8   &\textbf{81.2}  &83.4  &4.8 \tabularnewline
\hline 
Ours &\textbf{95.0}  &\textbf{88.6}  &\textbf{91.7}  &4.6   &\textbf{91.6}  &81.0  &\textbf{86.0} &4.8 \tabularnewline
\hline
\end{tabular}
\par\end{centering}
\end{table}

\subsection{Curved text}
\begin{figure}[!tp]
\begin{center}
\includegraphics[width=0.8\linewidth]{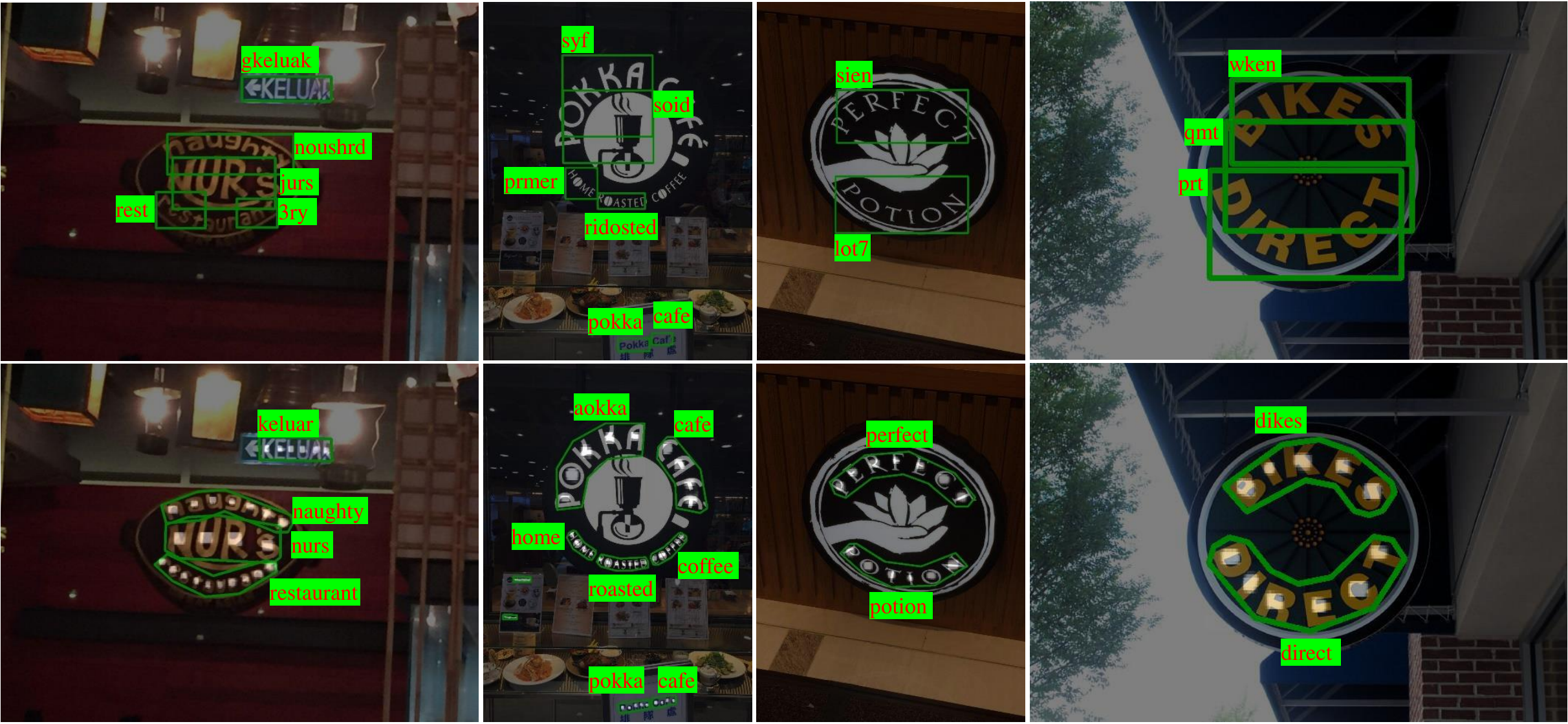}
\end{center}
\caption{Qualitative comparisons on Total-Text without lexicon. Top: results of TextBoxes~\cite{liao2017textboxes}; Bottom: results of ours.}
\label{fig:visu_compare}
\end{figure}

Detecting and recognizing arbitrary text (e.g. curved text) is a huge superiority of our method beyond other methods. We conduct experiments on Total-Text to verify the robustness of our method in detecting and recognizing curved text. Similarly, we input the test images with the short edges resized to 1000.  The evaluation protocol of detection is provided by \cite{CK2017}.
The evaluation protocol of end-to-end recognition follows ICDAR 2015 while changing the representation of polygons from four vertexes to an arbitrary number of vertexes in order to handle the polygons of arbitrary shapes.

\begin{table}[!tp]
\begin{centering}
\caption{Results on Total-Text. ``None" means recognition without any lexicon. ``Full" lexicon contains all words in test set.}
\label{tab_total}
\begin{tabular}{|c|c|c|c|c|c|}
\hline 
\multirow{2}{*}{Method} & \multicolumn{3}{c|}{Detection} & \multicolumn{2}{c|}{End-to-End}\tabularnewline
\cline{2-6} 
 & Precision & Recall & F-Measure & None & Full\tabularnewline

\hline 
\hline
Ch$^,$ng \emph{et al.} \cite{CK2017} &40.0 &33.0 &36.0 &-  & - \tabularnewline
\hline
Liao \emph{et al.} \cite{liao2017textboxes} &62.1 &45.5  &52.5 & 36.3 & 48.9 \tabularnewline
\hline 
\hline 
Ours &\textbf{69.0} &\textbf{55.0}  &\textbf{61.3} &\textbf{52.9} &\textbf{71.8} \tabularnewline
\hline 
\end{tabular}
\par\end{centering}
\end{table}

To compare with other methods, we also trained a model \cite{liao2017textboxes} using the code in \cite{liao2017textboxes} \footnote{https://github.com/MhLiao/TextBoxes} 
with the same training data. As shown in Fig.~\ref{fig:visu_compare}, our method has a large superiority on both detection and recognition for curved text. The results in Table~\ref{tab_total} show that our method exceeds \cite{liao2017textboxes} by $8.8$ points in detection and at least $16.6\%$ in end-to-end recognition. The significant improvements of detection mainly come from the more accurate localization outputs which encircle the text regions with polygons rather than the horizontal rectangles. Besides, our method is more suitable to handle sequences in 2-D space (such as curves), while the sequence recognition network used in \cite{liao2017textboxes,Li_2017_ICCV,Busta_2017_ICCV} are designed for 1-D sequences.

\subsection{Speed}

Compared to previous methods,  our proposed method exhibits a good speed-accuracy trade-off. It can run at 6.9 FPS with the input scale of $720 \times 1280$. Although a bit slower than the fastest method \cite{Busta_2017_ICCV},  it exceeds \cite{Busta_2017_ICCV} by a large margin in accuracy. Moreover, the speed of ours is about 4.4 times of \cite{Li_2017_ICCV} which is the current state-of-the-art on ICDAR2013.

\subsection{Ablation Experiments}
Some ablation experiments, including ``With or without character maps", ``With or without character annotation", and ``With or without weighted edit distance", are discussed in the Supplementary.

\section{Conclusion}
In this paper, we propose a text spotter, which detects and recognizes scene text in a unified network and can be trained end-to-end completely. Comparing with previous methods, our proposed network is very easy to train and has the ability to detect and recognize irregular text (e.g. curved text). The impressive  performances on all the datasets which  includes  horizontal text,  oriented text and curved text,  demonstrate the effectiveness and robustness of our method for text detection and end-to-end text recognition.

\section*{Acknowledgements}
This work was supported by National Key R\&D Program of China No. 2018YFB1\\004600, NSFC 61733007, and NSFC 61573160, to Dr. Xiang Bai by the National Program for Support of Top-notch Young Professionals and the Program for HUST Academic Frontier Youth Team.
%
%
%
\bibliographystyle{splncs04}
\bibliography{reference}

\begin{thebibliography}{10}
\providecommand{\url}[1]{\texttt{#1}}
\providecommand{\urlprefix}{URL }
\providecommand{\doi}[1]{https://doi.org/#1}

\bibitem{bengio2009curriculum}
Bengio, Y., Louradour, J., Collobert, R., Weston, J.: Curriculum learning. In:
  Proc. ICML. pp. 41--48 (2009)

\bibitem{bissacco2013photoocr}
Bissacco, A., Cummins, M., Netzer, Y., Neven, H.: Photoocr: Reading text in
  uncontrolled conditions. In: Proc. ICCV. pp. 785--792 (2013)

\bibitem{Busta_2017_ICCV}
Busta, M., Neumann, L., Matas, J.: Deep textspotter: An end-to-end trainable
  scene text localization and recognition framework. In: Proc. ICCV. pp.
  2223--2231 (2017)

\bibitem{CK2017}
Chng, C.K., Chan, C.S.: Total-text: {A} comprehensive dataset for scene text
  detection and recognition. In: Proc. ICDAR. pp. 935--942 (2017)

\bibitem{dai2016instance}
Dai, J., He, K., Li, Y., Ren, S., Sun, J.: Instance-sensitive fully
  convolutional networks. In: Proc. ECCV. pp. 534--549 (2016)

\bibitem{dai2016r}
Dai, J., Li, Y., He, K., Sun, J.: {R-FCN:} object detection via region-based
  fully convolutional networks. In: Proc. NIPS. pp. 379--387 (2016)

\bibitem{epshtein2010detecting}
Epshtein, B., Ofek, E., Wexler, Y.: Detecting text in natural scenes with
  stroke width transform. In: Proc. CVPR. pp. 2963--2970 (2010)

\bibitem{fastrcnn}
Girshick, R.B.: Fast {R-CNN}. In: Proc. ICCV. pp. 1440--1448 (2015)

\bibitem{girshick2014rich}
Girshick, R.B., Donahue, J., Darrell, T., Malik, J.: Rich feature hierarchies
  for accurate object detection and semantic segmentation. In: Proc. CVPR. pp.
  580--587 (2014)

\bibitem{gomez2017textproposals}
G{\'o}mez, L., Karatzas, D.: Textproposals: a text-specific selective search
  algorithm for word spotting in the wild. Pattern Recognition  \textbf{70},
  60--74 (2017)

\bibitem{graves2006connectionist}
Graves, A., Fern{\'{a}}ndez, S., Gomez, F.J., Schmidhuber, J.: Connectionist
  temporal classification: labelling unsegmented sequence data with recurrent
  neural networks. In: Proc. ICML. pp. 369--376 (2006)

\bibitem{SynthText}
Gupta, A., Vedaldi, A., Zisserman, A.: Synthetic data for text localisation in
  natural images. In: Proc. CVPR. pp. 2315--2324 (2016)

\bibitem{he2017mask}
He, K., Gkioxari, G., Doll{\'{a}}r, P., Girshick, R.B.: Mask {R-CNN}. In: Proc.
  ICCV. pp. 2980--2988 (2017)

\bibitem{resnet}
He, K., Zhang, X., Ren, S., Sun, J.: Deep residual learning for image
  recognition. In: Proc. CVPR. pp. 770--778 (2016)

\bibitem{he2017single}
He, P., Huang, W., He, T., Zhu, Q., Qiao, Y., Li, X.: Single shot text detector
  with regional attention. In: Proc. ICCV. pp. 3066--3074 (2017)

\bibitem{he2017deep}
He, W., Zhang, X., Yin, F., Liu, C.: Deep direct regression for multi-oriented
  scene text detection. In: Proc. ICCV. pp. 745--753 (2017)

\bibitem{lstm}
Hochreiter, S., Schmidhuber, J.: Long short-term memory. Neural Computation
  \textbf{9}(8),  1735--1780 (1997)

\bibitem{hu2017wordsup}
Hu, H., Zhang, C., Luo, Y., Wang, Y., Han, J., Ding, E.: Wordsup: Exploiting
  word annotations for character based text detection. In: Proc. ICCV. pp.
  4950--4959 (2017)

\bibitem{huang2014robust}
Huang, W., Qiao, Y., Tang, X.: Robust scene text detection with convolution
  neural network induced {MSER} trees. In: Proc. ECCV. pp. 497--511 (2014)

\bibitem{Jaderberg14c}
Jaderberg, M., Simonyan, K., Vedaldi, A., Zisserman, A.: Synthetic data and
  artificial neural networks for natural scene text recognition. CoRR
  \textbf{abs/1406.2227} (2014)

\bibitem{jaderberg2016reading}
Jaderberg, M., Simonyan, K., Vedaldi, A., Zisserman, A.: Reading text in the
  wild with convolutional neural networks. International Journal of Computer
  Vision  \textbf{116}(1),  1--20 (2016)

\bibitem{jaderberg2014deep}
Jaderberg, M., Vedaldi, A., Zisserman, A.: Deep features for text spotting. In:
  Proc. ECCV. pp. 512--528 (2014)

\bibitem{kang2014orientation}
Kang, L., Li, Y., Doermann, D.S.: Orientation robust text line detection in
  natural images. In: Proc. CVPR. pp. 4034--4041 (2014)

\bibitem{karatzas2015icdar}
Karatzas, D., Gomez{-}Bigorda, L., Nicolaou, A., Ghosh, S.K., Bagdanov, A.D.,
  Iwamura, M., Matas, J., Neumann, L., Chandrasekhar, V.R., Lu, S., Shafait,
  F., Uchida, S., Valveny, E.: {ICDAR} 2015 competition on robust reading. In:
  Proc. ICDAR. pp. 1156--1160 (2015)

\bibitem{karatzas2013icdar}
Karatzas, D., Shafait, F., Uchida, S., Iwamura, M., i~Bigorda, L.G., Mestre,
  S.R., Mas, J., Mota, D.F., Almaz{\'{a}}n, J., de~las Heras, L.: {ICDAR} 2013
  robust reading competition. In: Proc. ICDAR. pp. 1484--1493 (2013)

\bibitem{lee2016recursive}
Lee, C., Osindero, S.: Recursive recurrent nets with attention modeling for
  {OCR} in the wild. In: Proc. CVPR. pp. 2231--2239 (2016)

\bibitem{Li_2017_ICCV}
Li, H., Wang, P., Shen, C.: Towards end-to-end text spotting with convolutional
  recurrent neural networks. In: Proc. ICCV. pp. 5248--5256 (2017)

\bibitem{li2017fully}
Li, Y., Qi, H., Dai, J., Ji, X., Wei, Y.: Fully convolutional instance-aware
  semantic segmentation. In: Proc. CVPR. pp. 4438--4446 (2017)

\bibitem{liao2018textboxes++}
Liao, M., Shi, B., Bai, X.: Textboxes++: A single-shot oriented scene text
  detector. IEEE Transactions on Image Processing  \textbf{27}(8),  3676--3690
  (2018)

\bibitem{liao2017textboxes}
Liao, M., Shi, B., Bai, X., Wang, X., Liu, W.: Textboxes: {A} fast text
  detector with a single deep neural network. In: Proc. AAAI. pp. 4161--4167
  (2017)

\bibitem{liao2018rotation}
Liao, M., Zhu, Z., Shi, B., Xia, G.s., Bai, X.: Rotation-sensitive regression
  for oriented scene text detection. In: Proc. CVPR. pp. 5909--5918 (2018)

\bibitem{lin2017feature}
Lin, T., Doll{\'{a}}r, P., Girshick, R.B., He, K., Hariharan, B., Belongie,
  S.J.: Feature pyramid networks for object detection. In: Proc. CVPR. pp.
  936--944 (2017)

\bibitem{liu2016ssd}
Liu, W., Anguelov, D., Erhan, D., Szegedy, C., Reed, S.E., Fu, C., Berg, A.C.:
  {SSD:} single shot multibox detector. In: Proc. ECCV. pp. 21--37 (2016)

\bibitem{liu2017deep}
Liu, Y., Jin, L.: Deep matching prior network: Toward tighter multi-oriented
  text detection. In: Proc. CVPR. pp. 3454--3461 (2017)

\bibitem{lyu2018multi}
Lyu, P., Yao, C., Wu, W., Yan, S., Bai, X.: Multi-oriented scene text detection
  via corner localization and region segmentation. In: Proc. CVPR. pp.
  7553--7563 (2018)

\bibitem{neumann2010method}
Neumann, L., Matas, J.: A method for text localization and recognition in
  real-world images. In: Proc. ACCV. pp. 770--783 (2010)

\bibitem{neumann2012real}
Neumann, L., Matas, J.: Real-time scene text localization and recognition. In:
  Proc. CVPR. pp. 3538--3545 (2012)

\bibitem{neumann2016real}
Neumann, L., Matas, J.: Real-time lexicon-free scene text localization and
  recognition. {IEEE} Trans. Pattern Anal. Mach. Intell.  \textbf{38}(9),
  1872--1885 (2016)

\bibitem{redmon2016you}
Redmon, J., Divvala, S.K., Girshick, R.B., Farhadi, A.: You only look once:
  Unified, real-time object detection. In: Proc. CVPR. pp. 779--788 (2016)

\bibitem{ren2015faster}
Ren, S., He, K., Girshick, R.B., Sun, J.: Faster {R-CNN:} towards real-time
  object detection with region proposal networks. {IEEE} Trans. Pattern Anal.
  Mach. Intell.  \textbf{39}(6),  1137--1149 (2017)

\bibitem{risnumawan2014robust}
Risnumawan, A., Shivakumara, P., Chan, C.S., Tan, C.L.: A robust arbitrary text
  detection system for natural scene images. Expert Syst. Appl.
  \textbf{41}(18),  8027--8048 (2014)

\bibitem{long2015fully}
Shelhamer, E., Long, J., Darrell, T.: Fully convolutional networks for semantic
  segmentation. {IEEE} Trans. Pattern Anal. Mach. Intell.  \textbf{39}(4),
  640--651 (2017)

\bibitem{shi2017detecting}
Shi, B., Bai, X., Belongie, S.J.: Detecting oriented text in natural images by
  linking segments. In: Proc. CVPR. pp. 3482--3490 (2017)

\bibitem{shi2017end}
Shi, B., Bai, X., Yao, C.: An end-to-end trainable neural network for
  image-based sequence recognition and its application to scene text
  recognition. {IEEE} Trans. Pattern Anal. Mach. Intell.  \textbf{39}(11),
  2298--2304 (2017)

\bibitem{shi2016robust}
Shi, B., Wang, X., Lyu, P., Yao, C., Bai, X.: Robust scene text recognition
  with automatic rectification. In: Proc. CVPR. pp. 4168--4176 (2016)

\bibitem{shi2018aster}
Shi, B., Yang, M., Wang, X., Lyu, P., Yao, C., Bai, X.: {ASTER}: An attentional
  scene text recognizer with flexible rectification. IEEE transactions on
  pattern analysis and machine intelligence  (2018)

\bibitem{tian2015text}
Tian, S., Pan, Y., Huang, C., Lu, S., Yu, K., Tan, C.L.: Text flow: {A} unified
  text detection system in natural scene images. In: Proc. ICCV. pp. 4651--4659
  (2015)

\bibitem{tian2016detecting}
Tian, Z., Huang, W., He, T., He, P., Qiao, Y.: Detecting text in natural image
  with connectionist text proposal network. In: Proc. ECCV. pp. 56--72 (2016)

\bibitem{wang2011end}
Wang, K., Babenko, B., Belongie, S.: End-to-end scene text recognition. In:
  Proc. ICCV. pp. 1457--1464 (2011)

\bibitem{yao2012detecting}
Yao, C., Bai, X., Liu, Wenyu~and, M.Y., Tu, Z.: Detecting texts of arbitrary
  orientations in natural images. In: 2012 IEEE Conference on Computer Vision
  and Pattern Recognition. pp. 1083--1090. IEEE (2012)

\bibitem{yao2014unified}
Yao, C., Bai, X., Liu, W.: A unified framework for multioriented text detection
  and recognition. {IEEE} Trans. Image Processing  \textbf{23}(11),  4737--4749
  (2014)

\bibitem{yao2016scene}
Yao, C., Bai, X., Sang, N., Zhou, X., Zhou, S., Cao, Z.: Scene text detection
  via holistic, multi-channel prediction. CoRR  \textbf{abs/1606.09002} (2016)

\bibitem{yao2014strokelets}
Yao, C., Bai, X., Shi, B., Liu, W.: Strokelets: A learned multi-scale
  representation for scene text recognition. In: Proceedings of the IEEE
  Conference on Computer Vision and Pattern Recognition. pp. 4042--4049 (2014)

\bibitem{zhang2015symmetry}
Zhang, Z., Shen, W., Yao, C., Bai, X.: Symmetry-based text line detection in
  natural scenes. In: Proc. CVPR. pp. 2558--2567 (2015)

\bibitem{zhang2016multi}
Zhang, Z., Zhang, C., Shen, W., Yao, C., Liu, W., Bai, X.: Multi-oriented text
  detection with fully convolutional networks. In: Proc. CVPR. pp. 4159--4167
  (2016)

\bibitem{zhong2016deeptext}
Zhong, Z., Jin, L., Zhang, S., Feng, Z.: Deeptext: {A} unified framework for
  text proposal generation and text detection in natural images. CoRR
  \textbf{abs/1605.07314} (2016)

\bibitem{zhou2017east}
Zhou, X., Yao, C., Wen, H., Wang, Y., Zhou, S., He, W., Liang, J.: {EAST:} an
  efficient and accurate scene text detector. In: Proc. CVPR. pp. 2642--2651
  (2017)

\bibitem{zhu2018cascaded}
Zhu, Y., Liao, M., Yang, M., Liu, W.: Cascaded segmentation-detection networks
  for text-based traffic sign detection. {IEEE} Trans. Intelligent
  Transportation Systems  \textbf{19}(1),  209--219 (2018)

\bibitem{zhu2016scene}
Zhu, Y., Yao, C., Bai, X.: Scene text detection and recognition: Recent
  advances and future trends. Frontiers of Computer Science  \textbf{10}(1),
  19--36 (2016)

\bibitem{zitnick2014edge}
Zitnick, C.L., Doll{\'{a}}r, P.: Edge boxes: Locating object proposals from
  edges. In: Proc. ECCV. pp. 391--405 (2014)

\end{thebibliography}
%





\title{Supplementary of Mask TextSpotter}
\titlerunning{Mask TextSpotter}
%
\author{Pengyuan Lyu\thanks{Authors contribute equally.}\inst{1} \and
Minghui Liao\printfnsymbol{1}\inst{1} \and
Cong Yao\inst{2} \and
Wenhao Wu\inst{2} \and \\
Xiang Bai\thanks{Corresponding author.}\inst{1}}
%
\authorrunning{Pengyuan Lyu, Minghui Liao, Cong Yao, Wenhao Wu, Xiang Bai}
%

\institute{
Huazhong University of Science and Technology \and
Megvii (Face++) Technology Inc. \\
\email{lvpyuan@gmail.com, mhliao@hust.edu.cn, yaocong2010@gmail.com, wwh@megvii.com, xbai@hust.edu.cn}
}
\maketitle
\section{Ablation Experiments}
\begin{table}
\begin{centering}
\caption{Ablation experimental results. ``Ours (a)":  without character annotations from the real images; ``Ours (b)": without weighted edit distance.}
\label{tab_discussion}
\begin{tabular}{|c|c|c|c|c|c|c|c|c|c|c|c|c|}
\hline 
\multirow{3}{*}{Method} & \multicolumn{6}{c|}{ICDAR2013} & \multicolumn{6}{c|}{ICDAR2015}\tabularnewline
\cline{2-13} 
 & \multicolumn{3}{c|}{Word Spotting} & \multicolumn{3}{c|}{End-to-End} & \multicolumn{3}{c|}{Word Spotting} & \multicolumn{3}{c|}{End-to-End}\tabularnewline
\cline{2-13} 
 & S & W & G & S & W & G & S & W & G & S & W & G\tabularnewline
\hline 
Ours(a) &91.8 &90.3 &85.9 &90.7 &89.4 &84.6 &76.9 &71.6 &61.6 &76.6 &69.9 &59.8 \tabularnewline
\hline 
Ours(b) &91.4  &90.5  &84.3  &91.3  &89.9  &83.8  &75.9  &67.5  &56.8  &76.1  &67.1  &56.7 \tabularnewline
\hline 
Ours &\textbf{92.5}  &\textbf{92.0}  &\textbf{88.2} &\textbf{92.2}  &\textbf{91.1}  &\textbf{86.5}  &\textbf{79.3}  &\textbf{74.5}  &\textbf{64.2}  &\textbf{79.3}  &\textbf{73.0}  &\textbf{62.4} \tabularnewline
\hline 
\end{tabular}
\par\end{centering}
\end{table}

\noindent\textbf{With or without character maps} We train a model named ``Ours(det only)" which removes the subnet of the character maps from the original network to explore the effect of training detection and recognition jointly. As shown in Table 3 in the paper, the detection results of ``Ours" exceed ``Ours(det only)" by $0.7\%$ and $2.6\%$ on ICDAR2013 and ICDAR2015 respectively,  which demonstrate that the detection task can  benefit from the recognition task when jointly training.
\\
\textbf{With or without real-world character annotation} The experiment without real-world character annotations is also conducted. As shown in Table~\ref{tab_discussion},  although ``Ours(a)" is trained without any real-world character annotation, it still achieves competitive performances. More specifically, for horizontal text (ICDAR2013), it decrease ``Ours", which is trained with a few real-world character annotations, by  $0.7\%-2.3\%$ on various settings; on ICDAR2015, ``Ours(a)" still outperforms all other previous methods by a large margin.
\\
\textbf{With or without weighted edit distance} We conduct experiments to verify the effectiveness of our proposed weighted edit distance. The method of using original edit distance is named ``Ours(b)" and the results are shown in Table~\ref{tab_discussion}. As shown, weighted edit distance can boost the performance by at most 7.4 points of all experiments. 
\end{document}